\documentclass[12pt]{article}

\usepackage{mathptmx}
\usepackage{graphicx}

\title{Multi-scale metrics and self-organizing maps: a computational approach to the structure of sensory maps}

\title{Multi-scale metrics and self-organizing maps: a computational approach to the structure of sensory maps}

\author{
\\William H. Wilson\\
\normalsize{School of Computer Science and Engineering}\\
\normalsize{The University of New South Wales, Australia}\\ 
}

\date{}                                           

\begin{document}
\maketitle

\begin{abstract}
This paper introduces the concept of a bi-scale metric for use in the cooperative phase of the self-organizing map (SOM) algorithm. Use of a bi-scale metric allows segmentation of the map into a number of regions, corresponding to anticipated cluster structure in the data. Such a situation occurs, for example, in the somatotopic maps which inspired the SOM algorithm, where clusters of data may correspond to body surface regions whose general structure is known. When a bi-scale metric is appropriately applied, issues with map neurons that are not activated by any point in the training data are reduced or eliminated. The paper also presents results of simulation studies on the plasticity of bi-scale metric maps when they are retrained after loss of groups of map neurons or after changes in training data (such as would occur in a somatotopic map when a body surface region like a finger is lost/removed). The paper further considers situations where tri-scale metrics may be useful, and an alternative approach suggested by neurobiology, where some map regions adapt more slowly to stimuli because they have a lower learning rate parameter.
\end{abstract}


\section{Introduction}

This paper draws on material from neurobiology, self-organizing map neural networks, and the mathematical notion of a metric, so we begin by briefly describing relevant work and concepts from these three areas. We begin with biological sensory maps in general, and early experiments on their plasticity done by Kaas, Merzenich, et al. in the 1980s, then outline aspects of the Self-Organizing Map (SOM) algorithm developed by Kohonen, and next outline some concepts relating to the mathematical notion of a metric. In presenting the material on SOMs and metrics, we will introduce some notation that will be used in later sections of the paper.

\subsection{Sensory Maps}
It has been known since the 1930s that, to a large extent, adjacent sensory neurons of mammalian and other species are mapped to adjacent locations in the cerebral cortex. For example, the regions of cortex representing the index, middle, ring and little fingers are adjacent in that order in the somatosensory cortex. In this sense, the cortical somatosensory representation is a \emph{map} of the body surface from which the sensation derives. See for example \cite{Penfield1957}.

The adjacency is represented at levels of greater detail than that indicated in the fingers example above: for example, the two-dimensional array of sensory whiskers (``vibrissae'') in a mouse maps to an isomorphic two-dimensional array of regions in the barrel cortex of the mouse \cite{Woolsey1970}. Similarly, motor cortex regions responsible for controlling adjacent parts of the body are largely adjacent. Similar phenomena -- retinotopy and tonotopy -- apply to the cortical representations of sensory inputs from the eye and ear. Tone / frequency is 1-dimensional, so tonotopic maps are linear structures on the cortical surface (see e.g. \cite{Langers06102011}), whereas somatotopic maps and retinotopic maps are 2-dimensional (though multiple such maps may be used to build mental models of the 3-dimensional world, as in binocular vision in the case of humans).

Important sensory regions, such as the mouth and hands in a somatosensory map, and the foveal region in a retinotopic map, occupy disproportionately large amounts of the cortical map. Multiple maps exist for related senses, e.g. for proprioception and for light touch \cite{Kaas2013}. Vibrissae and barrel cortex are particularly interesting for the bi-scale-metric SOM algorithm variant introduced in this paper, since barrel cortex, like the bi-scale-metric system, divides the map into subregions in a fairly regular way. 

Consider, at the neural level, how such cortical maps might originate during brain development. Bundles of somatosensory axons come from a body part such as the hand, via spinal interneurons, and while some degree of adjacency may be preserved during this journey, neural circuits and cortical map are not completely hard-wired during development - see  the remarks below on mouse barrel cortex development \cite{vanderLoos1973}. Even if this part of the cortex were hard-wired during development, it would then still be subject to change via learning processes; see the remarks about plasticity, below. Hebb's principle \cite{Hebb1949}, together with the fact that tactile stimulation will usually cause a \emph{group} of adjacent sensory receptors to fire, suggests that somatosensory cortical neurons representing adjacent sensory receptors will be \emph{connected}, but not why they are physically \emph{adjacent}. The computational algorithm of \cite{Drake2004}, which infers the topological relationships of sensors from correlations between the sensor values over time, shows from a theoretical perspective that this sort of approach to development could work.

\subsection{Plasticity}
Somatotopic maps arise during brain development, with significant parts of map creation in mammals occurring after birth, as shown by \cite{vanderLoos1973} who carefully damaged a sensory whisker in mice at birth and found that the corresponding "barrel" in the mouse's cortex did not develop. In a stable environment, after cortical development, there is little reason for the somatosensory maps to change markedly with time, since the sensory \emph{structure} of the brain and body is reasonably stable -- although of course both body and brain will grow through childhood, and indeed post-puberty, e.g. \cite{Sowell2001, ReidEtAl2017a, Gould1999}.

However, after cortical development, if there is significant damage either to the sensory receptors, for example the loss of an eye, limb, or digit, or to the corresponding part of the brain, for example from a stroke, then a question arises about what happens to the somatosensory map. Similar considerations apply to motor maps. 

In the late 1970s and the 1980s, advances in single-neuron recording technology made it possible to explore what happens in such cases. In particular, \cite{Sur1979, KaasEtAl1981} demonstrated using a macaque monkey that if a finger is amputated, then the part of its somatosensory cortex that previously responded to sensations from the missing finger will over a period of time change so that it begins to respond to sensations from the adjacent fingers. Subsequently it became clear that cortical map reorganization can occur, to some extent at least, with stroke damage to parts of the cortex, \cite{MurphyCorbett2009}
and that, even in the absence of damage to cortex or sensory receptors, a similar reorganization takes place with players of stringed musical instruments: these musicians need more sensitivity in the hand that fingers the strings, and the cortex reorganizes to provide more ``coverage'' for those fingers \cite{Elbert1995}, with larger changes for those who started playing at an earlier age.

These phenomena are referred to as cortical map plasticity (e.g. \cite{XerriMerzenichEtAl1998}). Such plasticity is exhibited by brain maps, but not by the usual formulation of the artificial neural networks called self-organizing maps, to be described next.

\subsection{Self-Organizing Maps}
\cite{Kohonen1982, Kohonen2001} described a class of artificial neural networks called Self-Organizing Maps (SOM). SOMs are an artificial neural network model inspired by somatosensory maps, in which the simulated cortical neurons, often termed ``map nodes'', are laid out in a grid, often a 2-dimensional grid, sometimes 1-dimensional, and potentially of higher dimension, or cylindrical, toroidal, etc.), and in which spatially-related or otherwise similar inputs are mapped to neighboring nodes in the grid. SOMs have been extensively applied (see e.g. \cite{KohonenEtAl1996}) to non-neural data as a way of exploring the structure of that data. 
The SOM algorithm uses an iterative procedure, one version of which \cite[p 111]{Kohonen2001}, computes a \emph{neighborhood function}
$h_{j,i} = \mbox{exp}(-d(j, i)^2/2\sigma^2)$ for nodes $i, j$, where $d(j, i)$ is the distance between the nodes, and $\sigma$ is called neighborhood width. (The other version computes
a \emph{neighborhood set}, and will be discussed later.)

At each step, a winning node is chosen, rather like a single neuron firing, and changes made to the node connections, or weights, depend on distance from the winning node, the current neighborhood width, and the learning rate. Two parameters, neighborhood width $\sigma$, mentioned above, and a learning rate $\eta$, are systematically decreased over simulated time. When the process has converged, the result can be visualized, in the case of 2-dimensional input vectors and a 2-dimensional map, as a system in which the map nodes have distributed themselves among the input vectors.

There is, of course, more to the biological somatosensory map than is captured in standard SOMs (and Kohonen was well aware of this \cite[chapter 2, chapter 4]{Kohonen2001}, \cite{Kohonen1999}). In particular, first, the data input to the SOM algorithm are encoded in a way equivalent to assuming that the locations or spatial relationships of tactile neurons on the body surface giving rise to signals in the brain are known to the brain. Drake's algorithm \cite{Drake2004}, mentioned above, can address this issue; it can assemble, for example, an adjacency relationship between pixels in a sequence of systematically scrambled images using information about which pixels change together in the sequence. The pixels can be replaced by the use of arbitrary signals - sensor values, for example.

In a neural context, this would correspond to using neural firing sequence or co-occurence information to determine which sensory neurons were in proximity. 

\begin{figure}
\begin{center}
\includegraphics[width=9cm]{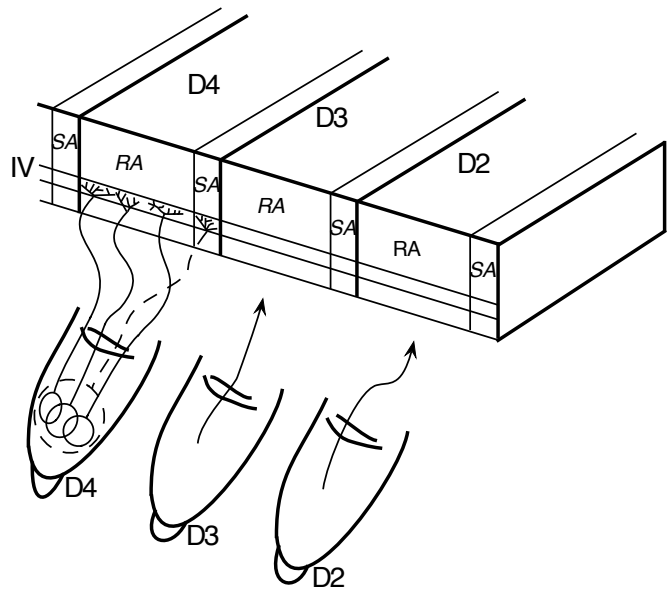}
\end{center}
\caption{\label{KaasFig8}Modular distribution of slowly-adapting (SA) and rapidly-adapting (RA) inputs in the region of cortex
representing the digits in area 3b of macaque monkeys. Each digit is represented by a central core of RA input
and flanked on one or both sides by sidebands of SA input. Other configurations are also possible. (Based on
\cite{KaasEtAl1981}, with permission.)}
\end{figure}

Second, in the standard SOM model, nodes are homogeneous. In comparison, in experiments with macaque monkeys, in somatosensory cortex, neurons in area 3b were found to be inhomogeneous, having been characterized as being activated by slowly adapting (SA) or rapidly adapting (RA) inputs \cite{Sur1979, KaasEtAl1981}.
Cortical neurons are of course also inhomogeneous in other ways, and can be classified based on e.g. dendritic architecture, axon length, and neurotransmitter used: SOMs don't attempt to capture these, either. 
In the regions representing the digits of the monkey's hand, neurons activated by SA inputs were found at the margins between the neurons representing individual digits (Fig. \ref{KaasFig8}). The standard SOM algorithm offers no way of addressing this distinction between SA and RA inputs. This is reasonable, given that Kohonen seems to have had in mind machine learning applications for which the RA/SA distinction was not important. \cite{Kohonen1982, Kohonen2001} certainly describes the correspondence between neural maps and the SOM algorithm. A major aim of this paper is to address the modeling of the RA/SA inhomogeneity using homogeneous neuron-like units and specially designed metrics.


A third difference between biological neural maps and SOMs is that biological cortical maps continue to learn throughout life, when and if their inputs change, or the cortex changes due to injury/illness, whereas the standard SOM algorithm converges once and then no longer learns, because the neighborhood function $h_{j,i} \rightarrow 0$ as neighborhood width $\sigma \rightarrow 0$ (and because the learning rate $\eta \rightarrow 0$ at the same time). Later parts of this paper are concerned with re-training a SOM after either the input patterns change, or a part of the map is ``lesioned'', though we will not try to address definitively the issue of biologically-plausible continual training for SOMs.

Fourth, inputs to biological cortical maps typically happen in such a way that several or many map neurons fire/win at or about the same time, since sensory stimulation, e.g. of the skin, will usually affect an area of skin, not just a single point. The SOM algorithm, in any particular time step, designates a single map neuron as a winner; in this respect, it is an essentially sequential algorithm. \cite[ch4]{Kohonen2001} considers biological issues including parallelism, and others, such as \cite{LawrenceEtAl1999}, have produced parallel versions of the SOM algorithm, which seem to be more for speed-up than for biological realism: for example, these approaches may use a batch version of the algorithm, computing the winners and weight changes for all input patterns in parallel, or may use a preliminary calculation to establish what the broad input data clusters are and then restrict changes for a map weight vector to the subset of data that belongs to the cluster associated with that map unit. This approach, while useful for its intended applications, does not address the kind of parallelism mentioned above.

We will not try to address the parallelism issue either, though again Drake's algorithm \cite{Drake2004} is somewhat relevant, in that it relies on temporal alignment of signals over a period of time to determine proximity or relatedness of the signal sources, which might be sensory neurons, or as Drake notes, stock market indicators, or weather measurements or other sensor signals from a range of locations.

Further differences exist: e.g. in biological neural networks, the number of inputs (i.e. the dimension of the input space) can change: \cite[p 125]{Kohonen2001} discusses how to deal with this in the SOM context.

\subsection{Metrics and Pseudometrics}
There is no general proof that the SOM procedure always converges, though special cases have been dealt with. Many studies of the convergence and convergence speed properties of SOMs exist; earlier ones are can be found in the extensive reference list in \cite{Kohonen2001}. Messages from these studies include that at least in the right circumstances, the map converges to the probability distribution of the training data as the epoch number $n \rightarrow \infty$ (e.g. \cite{YinAllison1995}).
Numerous variations are possible, including using triangular map grids, grids with extra connectivity like those described in \cite{jiangberryschoenauer2009}, and discretizations based on non-Euclidean spaces, such as those described in \cite{Ritter1999, Mayer-Neurocomputing2002}. 

The convergence arguments that exist are often phrased in terms of a particular kind of \emph{metric}. A metric, see e.g. \cite[p. 4]{Kohonen2001}, is a mathematical generalization of the notion of distance: it is a function $d(x,y)$  where $x$, $y$ are points (often in some vector space) and $d(x,y)$ is the real-valued distance between them, satisfying: 

\begin{enumerate}
\item $d(x,y) \ge 0$ and $d(x,y) = 0 \Leftrightarrow x = y$;
\item (symmetry) $\forall x, y \in X, d(x, y) = d(y, x)$; and
\item (triangle inequality): $\forall x, y, z \in X, d(x, z) \le d(x, y) + d(y, z)$.
\end{enumerate}

This does not capture all of the intuitive properties of a distance measure; for example, the \emph{discrete metric}, defined by $d(x, y) = 1$ if $x \ne y$ and $d(x, x) = 0$ satisfies the 3 metric axioms, above, but does not allow some pairs of points to be further apart than others, which is a normal part of the idea of distance.

If we weaken (i) by removing the requirement that $d(x,y) = 0 \Rightarrow x = y$, so that distinct points can be at zero distance from each other, we get what is called a \emph{pseudometric}.

Notice that if $\mu \ge 0$ then the weighted sum $d(x, y) + \mu \cdot \psi(x,y)$ of a metric $d$ and a pseudometric $\psi$ is always a metric - conditions (ii) and (iii) and the first part of (i) are trivial to check, and the second part of (i) is easy too: if $d$ is a metric and $\psi$ is a pseudometric, and we define $s(x, y) = d(x, y) + \mu \cdot \psi(x,y)$, with $\mu > 0$, then if $s(x,y) = 0$ , since $\mu \cdot \psi(x, y) \ge 0$, it must be that  $d(x,y) = 0$, so $x = y$. \footnote{This simple result seems to be part of the folklore of the theory of metrics.}

The metrics used in the standard SOM algorithm are based on a particular \emph{norm}, the $l^2_m$-norm: $d(\mathbf{x}, \mathbf{y}) = ||\mathbf{x} - \mathbf{y}||$, where $\mathbf{x} = (x_i)$, $\mathbf{y} = (y_i)$ are $m$-dimensional vectors, and the norm of $\mathbf{x}: || \mathbf{x} ||$ is defined to be $\left( { \sum_{i} {x_i^2}} \right)^{0.5}$. Let us refer to this as the $l^2$ norm unless we need to emphasise the dimension $m$.

In the SOM algorithm, metrics are applied both in the competitive phase and in the cooperative phase of the algorithm. Metrics other than the $l^2$-norm-based one have been used: see e.g. \cite{PlonskiZaremba2012}, who used a different metric in the competitive phase. In the cooperative phase, the $l^2$-norm based metric is sometimes replaced by a semi-qualitative indicator of neighborhood, neighborhood sets. These are normally designed to have metric-like properties, and we shall re-visit this alternative after introducing our variant metric for use in the SOM cooperative phase.

SOM variants with non-standard metrics for competitive and/or cooperative phases don't always work. For example, the discrete metric, defined above, unsurprisingly wouldn't work in the competitive phase, as it would give no basis for choosing the weight vector closest to the data item being presented, and also doesn't work in the competitive phase, where it again doesn't distinguish between map nodes near and distant from the winning node. However, experiments by the author \cite{Wilson2017} indicate that SOM variants using metrics based on the $l^p$-norm, $p > 1$, in the competitive phase do work, where $|| \mathbf{x} ||^p = \left( { \sum_{i} {x_i^p}} \right)^{1/p}$.

\begin{figure}
\begin{center}
\includegraphics[width=7.5cm]{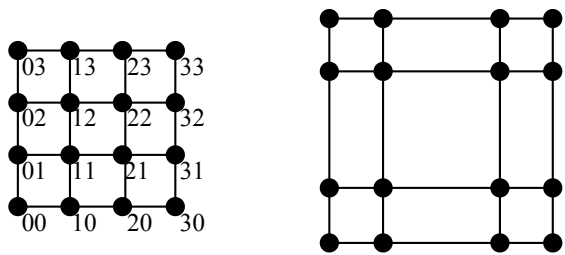}
\end{center}
\caption{\label{2x2metrics}Distances between points in the standard metric $d$ are shown on the left, for a $4 \times 4$ map. The map on the right is also $4 \times 4$, but shows distances using the metric $s(x, y)$ defined in the text. The vertical and horizontal lines between map units just indicate adjacency in the map. The pair of numerals beside each point indicates its grid coordinates.}
\end{figure}

The adaptation of the SOM algorithm that we shall describe uses a distance measure that is a positive-weighted sum of a metric and a pseudometric. The metric is the standard $l^2$-norm-based one; the pseudometric has \emph{groups} of vectors at distance zero from each other. Groups are assigned coordinates, and vectors in different groups are at a distance determined by a norm-based metric defined using the group coordinates, as exemplified below. An example of the distortion of distance achieved in this way is shown on the right in Fig. \ref{2x2metrics}, where the pseudometric divides the 16 vectors into 4 groups of 4. In this diagram, the metric $s$ defined by $s(x, y) = d(x, y) + \mu \cdot \psi(x,y)$, with $ \mu = 1$, and where $\psi(x, y) = 0$ between points within one of the four obvious clusters, while $\psi(x, y) = 1$ for points in horizontally or vertically adjacent clusters, and $\psi(x, y) = \sqrt 2$ for points in diagonally adjacent clusters.

In other words, we assign group coordinates $(0,0), (1,0), (0,1)$ and $(1,1)$ to the bottom-left, bottom-right, top-left, and top-right groups respectively, and define the pseudometric $s(x, y)$ to be the Euclidean distance between the group coordinates of $G_1$ and $G_2$, where $x \in$ group $G_1$ and $y \in$ group $G_2$. The distances in the left panel would be calculated as Euclidean distances using the coordinates shown. In the right panel, while the $s$-distances are approximately \emph{illustrated} in a Euclidean sense, clearly they cannot be calculated just by using the formula for Euclidean distance on the coordinates in Fig. \ref{2x2metrics}.

Having noted that there are two similar ways to look at this altered metric $s$ (i.e. using the formula $s = d + \mu \cdot p$, or using Euclidean distance on altered coordinates) we shall from now on use the formula $s = d + \mu \cdot \psi$.



Machine learning applications of the standard SOM algorithm work well in part because the number of map nodes used is typically small compared with the size of the set of data points whose structure the SOM learns. In the case of clustered data, in such a case the weight vectors for the map nodes tend to end up inside clusters. If the number of map nodes is increased, or if there are fewer map nodes than clusters in the data, the algorithm may place some map nodes between clusters of data points.

\begin{figure}
\begin{center}
\includegraphics[width=7.5cm]{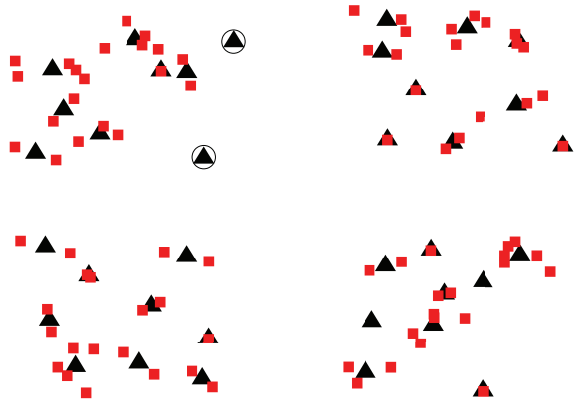}
\end{center}
\caption{\label{2x2clusters}Result of a standard SOM simulation with 4 clusters each of 20 data points, shown as squares, and a $6 \times 6$ map, with weight vectors shown as triangles. The circled weights lie between clusters.}
\end{figure}

In relation to the biological somatosensory maps that motivate SOMs, this would correspond to having neurons that respond to `stimuli' in the spaces between one's spread fingers, say, which of course doesn't make sense biologically. See Fig. \ref{2x2clusters}, which shows 4 clusters of pseudo-random data points (so 2-dimensional data) and superimposed on these, a 2-dimensional $6 \times 6$ SOM. Of the 36 map points, 34 lie within a cluster, while the remaining 2 lie between clusters. The number of between-clusters points is influenced by map structure: e.g. training using the same data, but a $7 \times 7$ map, so that the number of clusters does not divide evenly into the map dimensions, gave 11 between-clusters neurons.


\section{Variant Metrics and the SOM Algorithm}

With this background, we can state that the issues we try to address in this paper include the following:

\begin{itemize}
\item The standard SOM algorithm sometimes places map nodes outside of training data clusters. Our modified algorithm attempts to reduce this problem, in part for better biological plausibility.
\item With cortical maps, the incoming sensory data structure can change - either gradually (with change in activities or environment) or abruptly (due to trauma such as loss of a body part). We study how a modified SOM algorithm might respond to such situations.
\item With cortical maps, the map substrate can change, as with a stroke. We study how a modified SOM algorithm might respond to this sort of situation.
\item We compare direct simulation of SA/RA neurons (as in Fig. \ref{KaasFig8}) with the modified SOM algorithm.
\end{itemize}

The variant metrics to be described here are built on 2-dimensional maps, primarily because of the relationship with somatotopic maps, which are 2-dimensional, as they correspond primarily to sensor data from the 2-dimensional touch sensors on the body surface. However, there is no reason not to apply the same techniques to 1-dimensional maps and maps of dimension 3 or higher.

\subsection{Bi-scale metrics}
The modification of the SOM algorithm which we will now describe simply uses the metric $s$ defined above, in place of the standard one, in the SOM `cooperative process'. The SOM algorithm has three parts: the competitive, cooperative, and adaptive processes, and metric/norm-related computations occur in the competitive and cooperative processes.

In the standard SOM algorithm, the $l^2$-norm is the basis for the metric used in both the competitive and cooperative processes. In the competitive process, weight vectors are compared with input patterns in a space whose dimension is the number of input neurons.

In the cooperative process, pairs of map coordinates are compared with each other, in a space of dimension 2 if a 2-D map is being used. The metric used for comparison does not need to be the same in both processes; the point of the modifications described in this paper is to change the structure of the map, so the modified metric is applied only to comparisons of pairs of map coordinates, so only in the cooperative process.

Using the notation of \cite{Haykin2009}, a common version of the cooperative process defines a neighborhood function $h$ that depends on distance:

\begin{equation}
h_{j,i(\mathbf{x})}(n) = \mbox{exp}(-d(j, i(\mathbf{x}))^2/2\sigma_n^2)
\end{equation}

\noindent
where $i(\mathbf{x})$ is the winning input node for input pattern $\mathbf{x}$, $\sigma_n$ is the `neighborhood width' parameter, and $n$ is epoch number, and here $d$ will be replaced by the modified metric $s$. See also \cite[p. 111]{Kohonen2001}, which also describes an alternative version of $h$ based on ``neighborhood sets'' (see next paragraph).

This modified neighborhood function is then used in the usual adaptive process, which modifies the weight vector $\mathbf{w}_j$ between each map node $j$ and the input units in the standard way, i.e. according to $\mathbf{w}_j(n+1) = \mathbf{w}_j(n) + \eta_n h_{j,i(\mathbf{x})}(n) (\mathbf{x} - \mathbf{w}_j(n))$, where $\eta_n$ is the learning rate in epoch $n$. The adaptive process is thus affected just because $s$ replaces $d$ in the formula for $h_{j,i(\mathbf{x})}(n)$, which now becomes $\mbox{exp}(-s(j, i(\mathbf{x}))^2/2\sigma_n^2)$.

The neighborhood set version of the classic SOM algorithm cooperative process, mentioned earlier, that uses neighborhood sets in the cooperative phase, might possibly be adaptable to achieve a similar effect to bi-scale metrics. The neighborhood set method involves reducing the size of the neighborhood sets of each map node over time. One could modify this scheme by keeping track of the group that the map node is in, and the group(s) that its neighbors are in, and ``pruning'' the neighborhood set of each node to remove map nodes that are in different groups.

The details would be complicated by a likely need to maintain some connection between groups, which might mean that particularly in the early stages (i.e. for low epoch number $n$) it would be necessary to have some degree of neighborhood set membership for map nodes in adjacent groups. Thus there might need to be two or more stages: initially, nodes in other groups could belong to the neighborhood set of a map node; later, only nodes in the same group would be eligible to belong to the neighborhood set of a map node. These restrictions would be superimposed on the progression of neighborhood set size as $n$ increases that is normal for the neighborhood set method. We have not attempted to implement this.

\subsection{Tri-scale Metrics}
\begin{figure}
\begin{center}
\includegraphics[width=6.5cm]{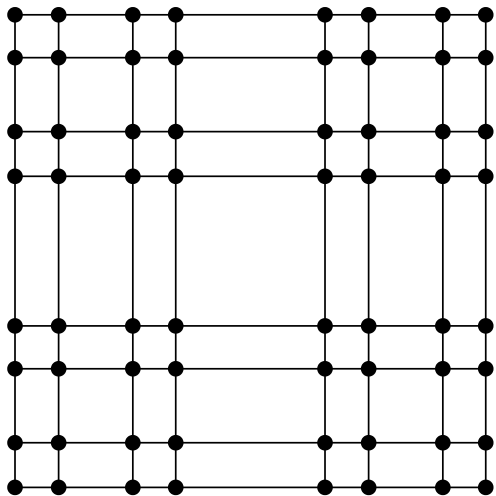}
\end{center}
\caption{\label{2x2x2clusters}Tri-scale metric, showing distances using the metric $s_2(x, y)$ defined in the text. The whole map is $8 \times 8$, the groups are $4 \times 4$, and the subgroups are $2 \times 2$.}
\end{figure}

Other unusual metrics could be used with the SOM algorithm. One example would be to extend the bi-scale metric idea to allow not only \emph{groups} of map units, where units are in the same group if they are at ``distance'' zero from each other according to the pseudometric $\psi$, but also \emph{subgroups} of map units, defined by another pseudometric $\psi_2$. $\psi_2$ is a refinement of $\psi$ in the following sense: if $\psi(x, y) = 0$, then $\psi_2(x, y) = 0$ too. Map neurons $x$ and $y$ are in the same subgroup if $\psi_2(x, y) = 0$, in which case obviously they are also in the same group. Thus subgroups nest within groups. The composite metric is, in this case, of the form $s_2(x, y) = d(x, y) + \mu \psi(x,y) + \lambda \psi_2(x, y)$. The effect on distances between map units is illustrated in Fig. \ref{2x2x2clusters}.


As with bi-scale metrics, tri-scale metrics might possibly be adaptable to the neighborhood set variant of the SOM cooperative phase, but it would be more complicated. Perhaps in a first stage, members of other groups could be in the neighborhood set of a map node; in a second stage, members of other groups would be excluded but members of subgroups of the same group as a map node could be in the neighborhood set of that map node; in a third stage, only members of the same subgroup could be members of the neighborhood set of the map node. Again, we have not tried to implement this scheme.

\section{SOM Simulations with Modified Metrics}

Simulation experiments using the metric $s$ show that it effectively deals with the problem of map neurons falling between clusters. Potentially, $\mu$ in the formula $s(x, y) = d(x, y) + \mu \cdot \psi(x, y)$ is a parameter that could be adjusted in these simulations, but in fact $\mu = 1$ has worked well, at least in the cases considered. Recomputing the metric involves calculating the coordinates for the map nodes in the $s$ metric, as follows: let $g$ (standing for $groupSideSize$ denote the number of rows/columns in a square group of map nodes (2-dimensional map) - so the groups contain $g \times g$ map nodes. Then the $\psi$-coordinates of a neuron whose normal coordinates are $(x, y)$, where coordinate value start at 0 (so the bottom-left node has the coordinates $(0, 0)$ as in Fig. \ref{2x2metrics}) will be $(x \ \mathbf{div} \  g, y \  \mathbf{div} \  g)$, where $\mathbf{div}$ denotes integer division. The value of $\psi(x, y)$ is then calculated as the Euclidean distance between the \emph{group} coordinates of $x$ and $y$, and then we are able to calculate $s(x, y)$.

The SOM simulations described in this paper started with pseudo-random initialization of weight vectors: other initializations might have led to better performance \cite[section 3.7]{Kohonen2001}, but pseudo-random initialization avoids biasing the algorithm or providing it with further hints about the data to be organized. In the case of multi-stage simulations, later stages start with the weight vectors as they were at the end of the previous stage.

Rectangular map layouts (as opposed to say hexagonal or irregular layouts) have been used for greater simplicity in splitting the map into groups (and/or subgroups).


\begin{figure}
\begin{center}
\includegraphics[width=8.5cm]{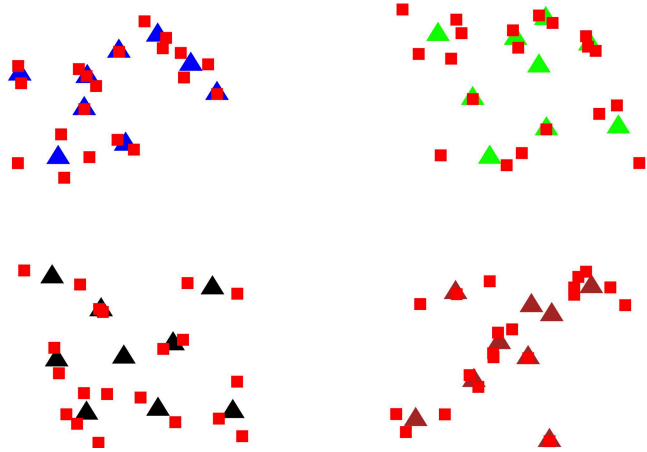}
\end{center}
\caption{\label{modifiedSOM6x6map_3colside}Results of simulation using modified SOM algorithm and a $6 \times 6$ map. Compare with the results using the standard-metric SOM algorithm in Fig. \ref{2x2clusters}, which had the same data points and map size. In this simulation, all weights lie inside data point clusters. As in Fig. \ref{2x2clusters}, squares represent data points, and triangles represent weight vectors. Different coloured triangles indicate different groups.}
\end{figure}

\subsection{Basic simulations using the metric s(x, y)}
Fig. \ref{modifiedSOM6x6map_3colside} shows results obtained with the modified metric $s$, for the same $2 \times 2$ clusters training data as in Fig. \ref{2x2clusters}, where the pseudometric was used to divide the map into four $3 \times 3$ sub-maps. (Fig. \ref{2x2clusters} used the standard SOM algorithm.) With the modified version of the algorithm, all weight vectors lie within the convex hull of one of the clusters of data. There are obvious limits to the use of this type of modified metric to ``clean up'' the map produced by the SOM algorithm. The group structure must correspond to the cluster structure in the data. This restriction should not be problematical for somatosensory and similar cortical maps, since the body/brain system during development has some knowledge of its own structure. It would be a problem with training data whose structure is totally unknown.

\begin{figure}
\begin{center}
\includegraphics[width=12cm]{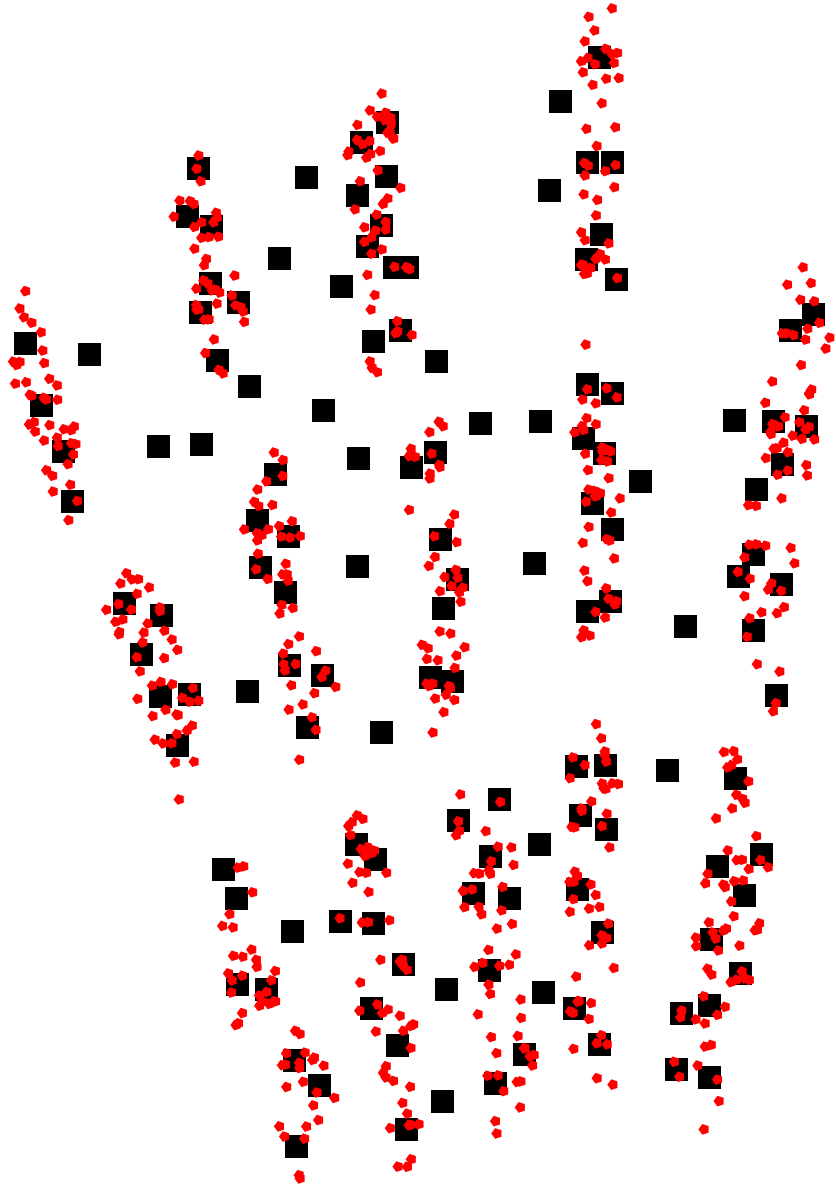}
\end{center}
\caption{\label{Hand15x9standardSOMe=5K}Results of simulation using randomly-generated ``hand'' data and the standard SOM algorithm with a $15 \times 9$ map. Compare with the results using the modified-metric SOM algorithm in the next figure. The data points are shown using pentagon symbols, the weight vectors for map neurons are shown as squares. The 27 weight vectors that are significantly outside of data clusters are circled; a couple of others are suspect.}
\end{figure}

\begin{figure}
\begin{center}
\includegraphics[width=12cm]{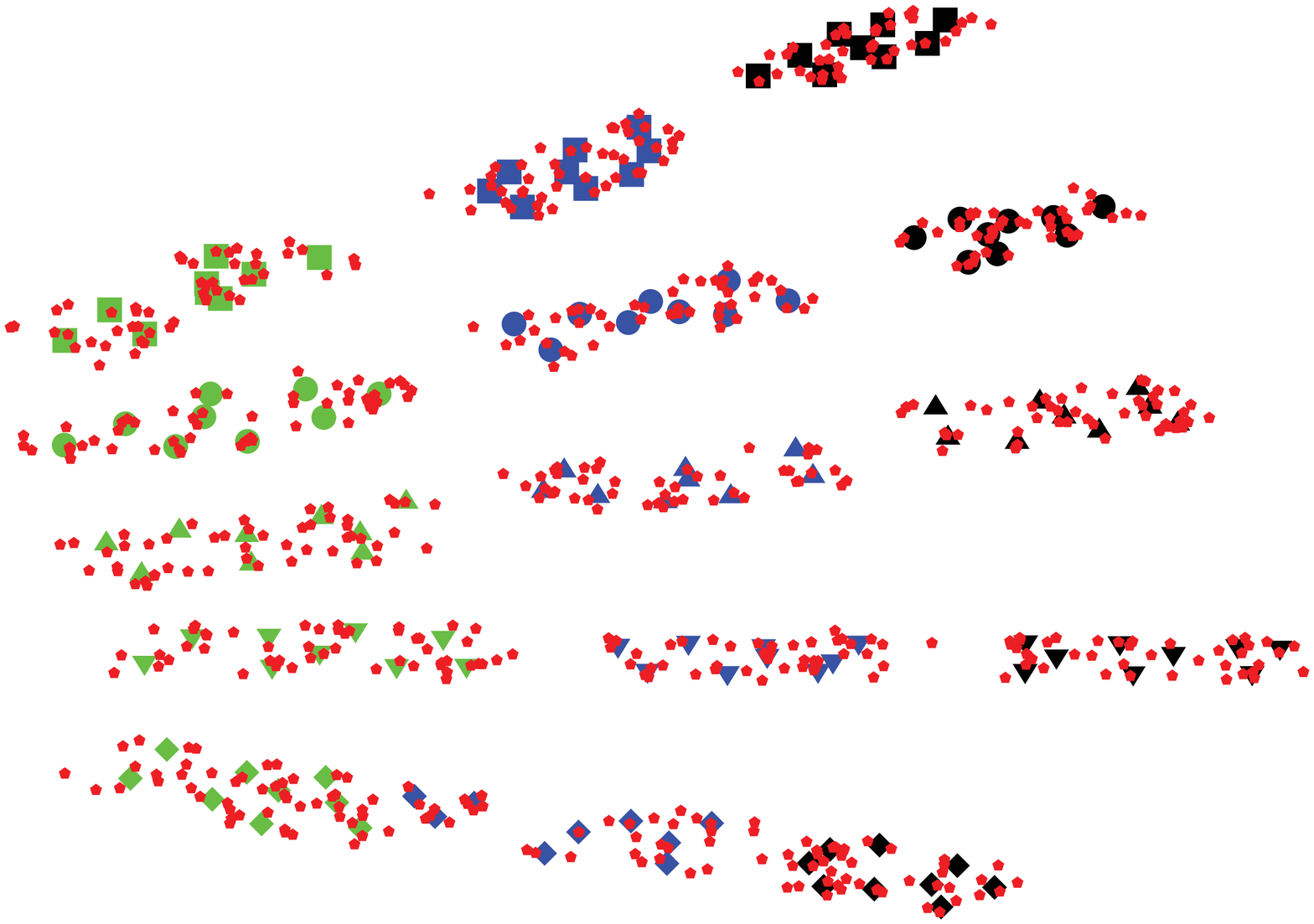}
\end{center}
\caption{\label{Hand15x9colside3modifiedSOMe=5K}In Figures \ref{Hand15x9colside3modifiedSOMe=5K}-\ref{hand-wholefinger-secondary}, data points are shown as pentagons and weights as other symbols, with a different shape + shade/color combination for each weight group. This figure shows the results of simulation using randomly-generated ``hand'' data and the \emph{modified} SOM algorithm with metric $s$, with a $15 \times 9$ map and weight vectors grouped into $3 \times 3$ groups, signified by different map symbols. Compare with the results using the standard SOM algorithm in the previous figure. All weight vectors lie inside the clusters of data points. The three $3 \times 3$ groups that correspond to the two ``thumb''  segments and are distributed within the data for these segments.}
\end{figure}

Figs. \ref{Hand15x9standardSOMe=5K} and \ref{Hand15x9colside3modifiedSOMe=5K} show typical results obtained with the standard and modified metrics respectively, but using data points inspired by the Kaas experiments \cite{KaasEtAl1981}. The 675 data points may be thought of as locations of sensory neurons on the digits of a hand, with 3 segments for each finger, and 2 segments for the thumb. (Note that while distinct biological \emph{digits} have no sensory neurons in the gaps between them, sensory neurons are continuous between the \emph{segments} of a digit, so this hand model is imperfect - though something like this would be a reasonable represention of the situation with sensory whiskers and barrel cortex.)

These data were generated as pseudo-random uniformly distributed points within rectangles bounding the segments of digits, with segments containing between 40 and 65 points. The map size could be viewed as   compression of this sensory data from the digits; obviously real sensory data would be more extensive, and the map size larger. In this case the pseudometric mechanism is used to subdivide the  $15 \times 9$ map into groups of size $3 \times 3$. It can be seen that in Fig. \ref{Hand15x9standardSOMe=5K} (standard SOM), a large number of map nodes lie outside any of the training data clusters, whereas in Fig. \ref{Hand15x9colside3modifiedSOMe=5K} (modified SOM), all of the map nodes lie inside the convex hull of a training data cluster, and for the four fingers, each map node group lies within a segment. Although neither node nor group coordinates are shown in Fig. \ref{Hand15x9colside3modifiedSOMe=5K}, adjacent groups do in fact inhabit adjacent segments. With the thumb training data, there are two segments but three map node groups, so these groups split across the two segments. (Figs. \ref{grid-with-gap} and \ref{hand-2-seg-thumb} illustrate a simulation where the map has only two thumb segment groups.)


\subsection{Simulations with virtual excision of digits}
The standard SOM algorithm does not consider situations where the data mix varies with time, e.g. by changing significantly after the SOM has converged. Convergence means no further changes can occur, since $\sigma$ and $\eta$ affect the size of weight changes, and both are close to 0 at the end of the simulation. This situation arises in attempts to simulate the macaque digit excision experiments of \cite{KaasEtAl1981}. A simple strategy, though one that introduces an extra issue in relation to biological plausibility, is to train the SOM using the original data mix, so that the learning rate $\eta \rightarrow 0$ and the neighborhood width parameter $\sigma \rightarrow 0$, then change the training patterns, e.g. to reflect simulated digit excision and, retaining the weights that have been learned, reset $\eta$ and $\sigma$ to higher values and resume the SOM learning processes. In a biological context, sleep might perform such a function, noting that somatotopic map modification appears to take place over a period of weeks or more \cite{KaasEtAl1981}, and motor skills, which may depend on a similar map, also seem to be consolidated in a fairly long time frame.

When training is resumed, replacing the original hand dataset with a dataset that is missing a finger (or with a dataset that is missing part of a finger), the opportunity arises to change some of the parameters of the SOM algorithm. For example, one might use a different initial value for the learning rate, $\eta(0)$, and/or neighborhood width, $\sigma(0)$. The decay factors for these could be changed, and a different number of training epochs could be used. The SOM algorithm in general turns out to be relatively insensitive to changes to $\eta(0)$ and its decay factor: even if no decay occurs, then provided $\eta(0)$ is ``reasonable'', a sensible  map is produced, although the map will not completely converge; instead when $\sigma \rightarrow 0$, each map neuron's weight vector will wander in a region bounded the data points for which it wins (modified by the frequency with which such data points are presented). In contrast, if $\sigma(0)$ is too small, or does not decay, then the map will not converge (cf. \cite[p111]{Kohonen2001}).

\begin{figure}
\begin{center}
\includegraphics[width=12cm]{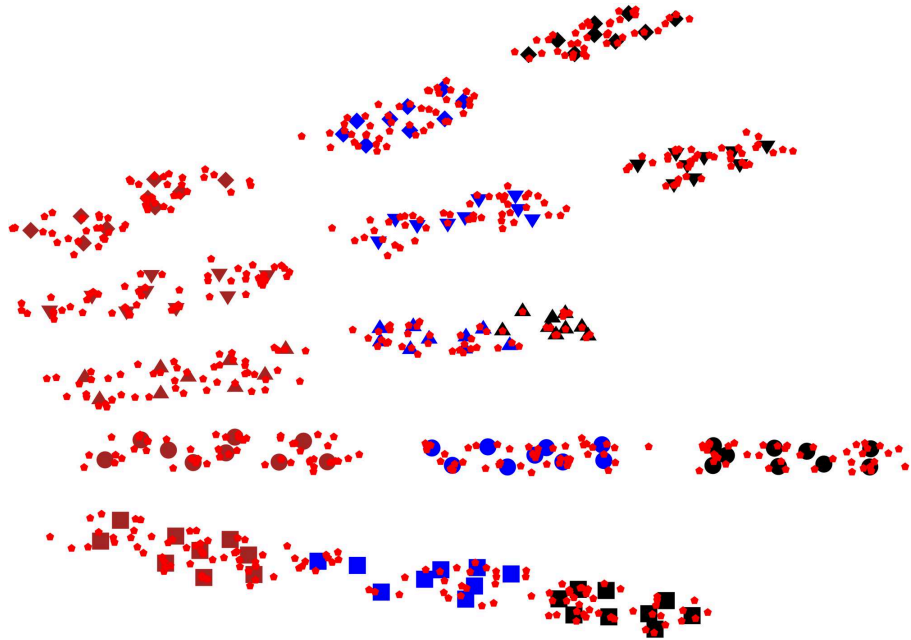}
\end{center}
\caption{\label{hand-amp1-secondary-adjustedplot}Results of two-stage simulation beginning with the randomly-generated ``hand'' data as in Figs. \ref{modifiedSOM6x6map_3colside}, \ref{Hand15x9standardSOMe=5K} with the \emph{modified} SOM algorithm with metric $s$, and a $15 \times 9$ map with weight vectors grouped into $3 \times 3$ groups. In the secondary training stage, the data points corresponding to the rightmost segment of the middle finger were removed, and the SOM process resumed with parameter adjustments as described in the text. The map units that were previously associated with the rightmost segment of the middle finger are now in the middle segment of the middle finger.}
\end{figure}

That said, experiments showed that setting $\sigma(0)$ to around 1 when training was resumed still resulted in a sensible map, though $\sigma(0) = 0.5$ resulted in problematical maps, at least for the datasets used. Recall that  $\sigma(0)$ is normally the radius of the map. Fig. \ref{hand-amp1-secondary-adjustedplot} shows the outcome of a simulation where the SOM was trained on the original data, then the data representing the final (i.e. rightmost) segment of the middle finger were removed, the learning rate $\eta(0)$ restored to its original value of 0.1, $\sigma(0)$ reset to 1 (not the radius of the map), and both parameters decayed over the epochs of the secondary training.

As can be seen in Fig. \ref{hand-amp1-secondary-adjustedplot}, the map weights previously associated with the missing finger segment have migrated back into the second segment of the middle finger. The weights previously associated with the second segments have moved over to ``make room'' for the displaced weights, but they remain within their original segment (compare the ``thumb'', where the three $3 \times 3$ groups of weights share the data for the two thumb segments in a different way, though this may be an artifact of the particular pseudo-randomly-generated data points that occur in the representation of thumb and middle finger).


\begin{figure}
\begin{center}
\includegraphics[width=12cm]{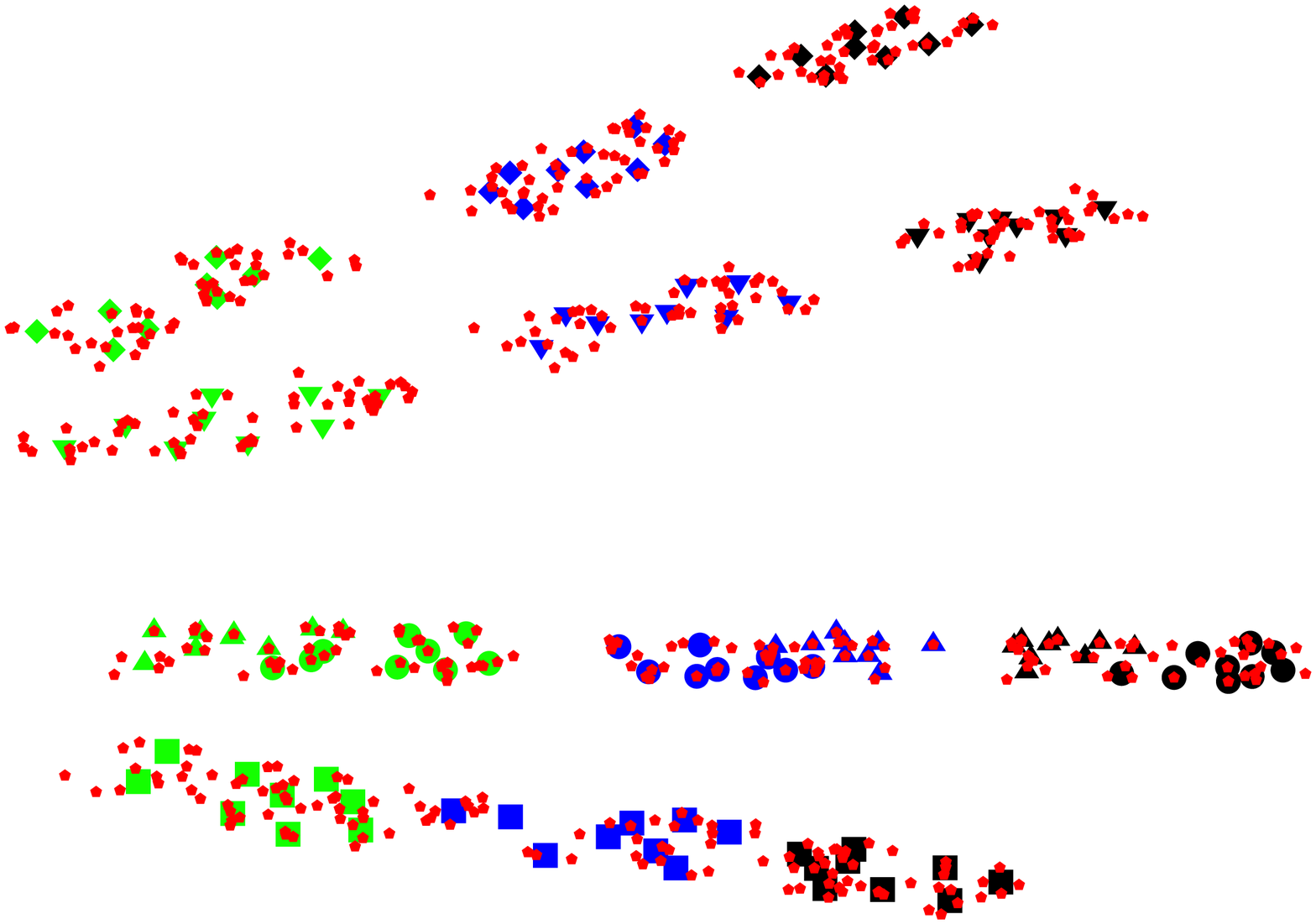}
\end{center}
\caption{\label{hand-wholefinger-secondary}Results of two-stage simulation begining with the randomly-generated ``hand'' data as in Figs. \ref{modifiedSOM6x6map_3colside}, \ref{Hand15x9standardSOMe=5K}  with the \emph{modified} SOM algorithm with metric $s$, and a $15 \times 9$ map with weight vectors grouped into $3 \times 3$ groups. In the secondary training stage, the data points corresponding to the whole of the middle finger were removed, and the SOM process resumed with adjusted parameters. The map units previously associated with the middle finger (triangles pointing up) have moved into the finger below (joining the weights plotted as circles).}
\end{figure}

Fig. \ref{hand-wholefinger-secondary} shows the outcome of a similar simulation where the data corresponding to the whole of the middle finger was removed for secondary training. In this case, the weights previously associated with the three segments of the middle finger have migrated into the corresponding segments of one of the adjacent fingers. So, after this retraining, all nodes continue to represent parts of some finger. This is broadly consistent with neurophysiological interpretations of phantom limb sensations (e.g. \cite{RamachandranHirstein1998}).


\subsection{Maps with Gaps}

\subsubsection{A ``hand'' map with two segments in the ``thumb''}
\begin{figure}
\begin{center}
\includegraphics[width=5cm]{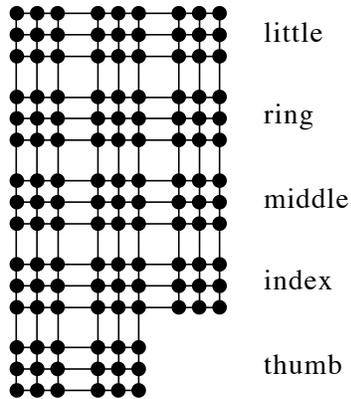}
\end{center}
\caption{\label{grid-with-gap} SOM map template with a gap at bottom right, corresponding to four fingers with three segments and one thumb with two segments.}
\end{figure}

\begin{figure}
\begin{center}
\includegraphics[width=12cm]{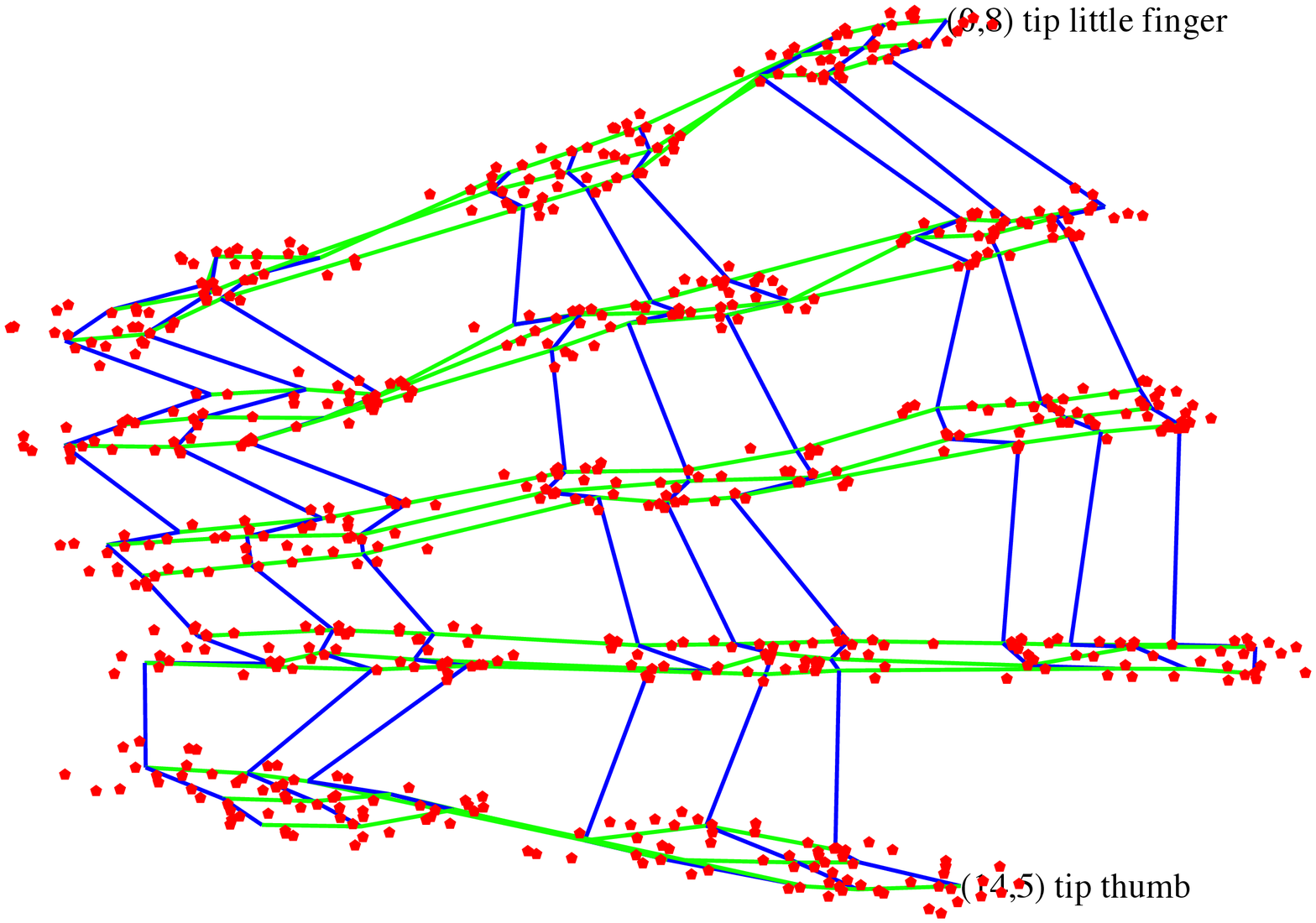}
\end{center}
\caption{\label{hand-2-seg-thumb} Result of simulation using the map from Fig. \ref{grid-with-gap} and the hand data seen in Fig. \ref{Hand15x9colside3modifiedSOMe=5K}, using the \emph{modified} algorithm with metric $s$. This diagram uses line segments to indicate adjacency relationships between map nodes, which occur at the junctions of the line segments. The coordinate labels indicate the positions of key map nodes.}
\end{figure}

As noted above in the text relating to Fig. \ref{Hand15x9colside3modifiedSOMe=5K}, the SOM topology used for the hand data had $5 \times 3$ groups (each of $3 \times 3$ units) even though the data had only 2 groups in the final or thumb row. It was interesting to see what the bi-scale SOM algorithm did with this scenario: in fact it squeezed three map node groups into the two thumb training data clusters/segments. It is also interesting to consider what happens if the map is modified so that there are only two groups of $3 \times 3$ units in the 5th row, as illustrated in Fig. \ref{grid-with-gap}.  This configuration could be described as $4 \times 3 + 1 \times 2$ groups of $3 \times 3$.  The rectangular SOM topology used previously, in e.g. the experiments summarized in Figs. \ref{Hand15x9standardSOMe=5K} and \ref{Hand15x9colside3modifiedSOMe=5K} is symmetric about vertical and horizontal axes, and the $(0,0)$-map unit can end up with weights lying in any of the four corner groups of data, depending on the pseudo-random initialization of the map weights, and this is not noticeable in the plots; one has to check the map neuron coordinates to detect this. However, the gapped map is no longer symmetric. When a SOM with such a map is simulated (with the modified, two-stage algorithm), only occasionally -- about one in four times -- will the map align with the underlying topology of the data. When it does align, the thumb data is inhabited by the weight vectors of the two final row $3 \times 3$ groups of map units, and they are adjacent to the first and second $3 \times 3$ groups of map units that inhabit the forefinger data, in a natural way, as illustrated in Fig. \ref{hand-2-seg-thumb}. For biological neural systems, this raises the issue of how developing brains align such information; the answer presumably relates to the adjacency of sensory input from the palm of the hand, which provides extra constraints. 


\subsubsection{A map with a gap in the middle: stroke simulation}
The map just discussed was designed to correspond better to the ``hand'' data that it was given to organize (given some information about the structure of the data). Experiments described earlier dealt with the situation where an \emph{input stimulus group} (i.e. training data cluster), which was present during initial training, disappears, as happens when part of a finger is removed.

\begin{figure}
\begin{center}
\includegraphics[width=6cm]{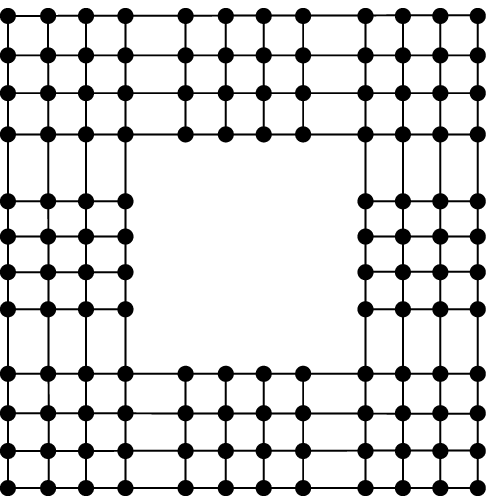}
\end{center}
\caption{\label{map-with-gap} $12 \times 12$ SOM map template configured as a $3 \times 3$ array of $4 \times 4$ groups, with a $4 \times 4$ gap in the middle. Map unit adjacency lines across the gap have been omitted to emphasize the gap.}
\end{figure}

Another variation involves deleting a group of \emph{map units}. This corresponds to what happens following a stroke: neurons that might belong to e.g. a somatotopic map are now dead or inoperative; what does (a two-stage version of) the modified SOM algorithm do in this situation? For this purpose we used as training data 9 clusters, each containing 80 uniform pseudo-random 2-dimensional vectors, and for the map, a $12 \times 12$ map organized into nine $4 \times 4$ groups, so arranged overall as a $3 \times 3$ array of groups, but with the center group of map nodes removed in the second stage of training. The map topology used in the second stage is shown in Fig. \ref{map-with-gap}.

\begin{figure}
\begin{center}
\includegraphics[width=12cm]{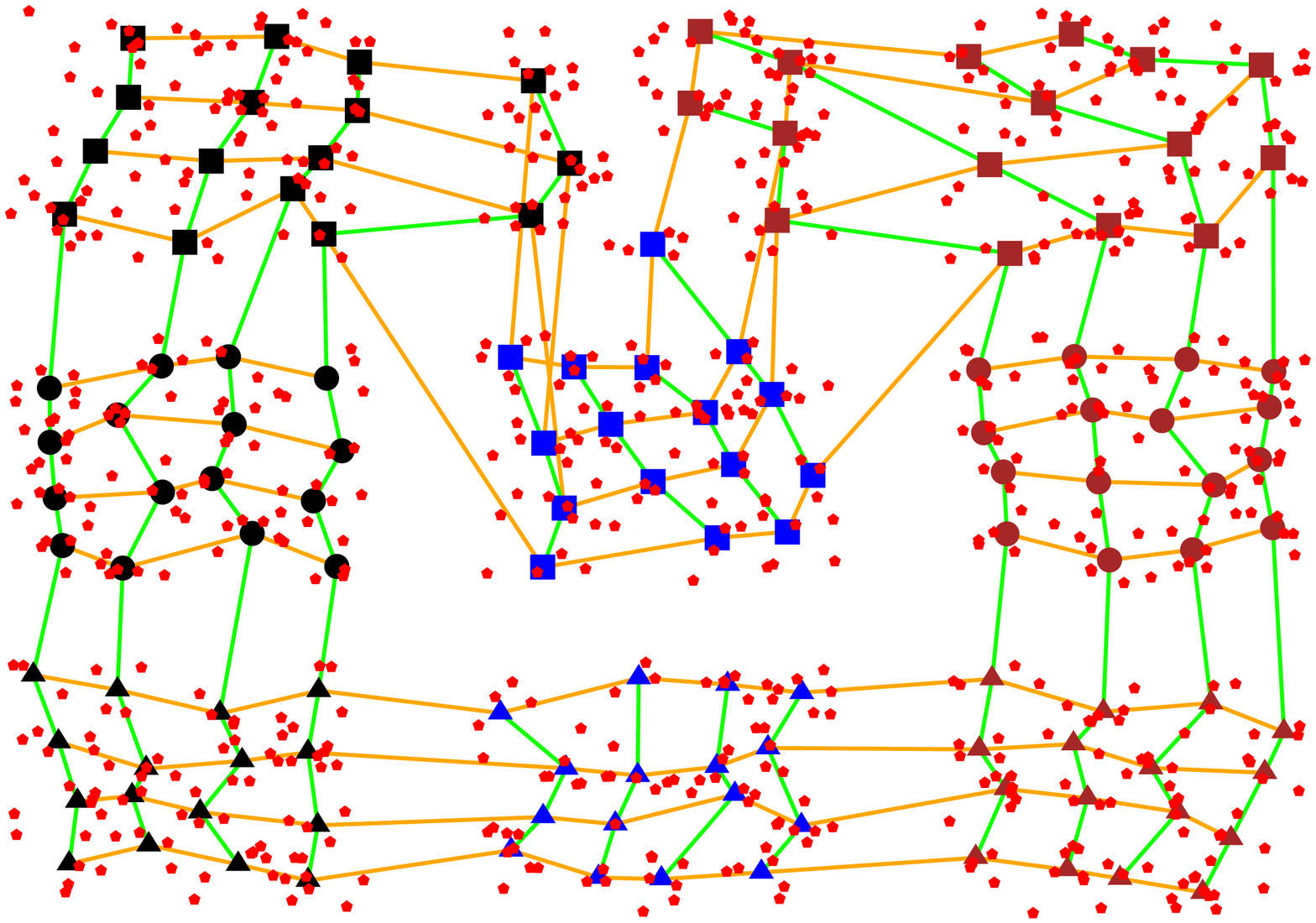}
\end{center}
\caption{\label{biscale-with-gap} Result of simulating a bi-scale metric SOM (metric $s$) with a ``stroke''-type gap in the middle of the map (that is, a group of map neurons is missing). All map neurons lie within data clusters, but the map seems subjectively strange (see text). This diagram indicates adjacency relationships between map nodes.}
\end{figure}

In the first stage of training, the behavior that we would expect from earlier simulations in this paper occurs: the 9 data clusters align with the 9 groups in the map. In stage 2, we are in effect inviting one or more of the subgroups to fragment in order to cover the data points whose group of map units were deleted between stages. The algorithm manages this with the modified metric, but, in the particular simulation illustrated in Fig. \ref{biscale-with-gap}, \emph{almost all} of the map neurons that previously served the data cluster in the left of center position now respond to data points in the central cluster, while the left of center cluster is now served by map neurons that have migrated from the top-left and top-right groups. From a biological realism point of view, this seems to be an extreme and rather strange reaction, with changes to the responses within 4 of the 9 clusters.


\begin{figure}
\begin{center}
\includegraphics[width=12cm]{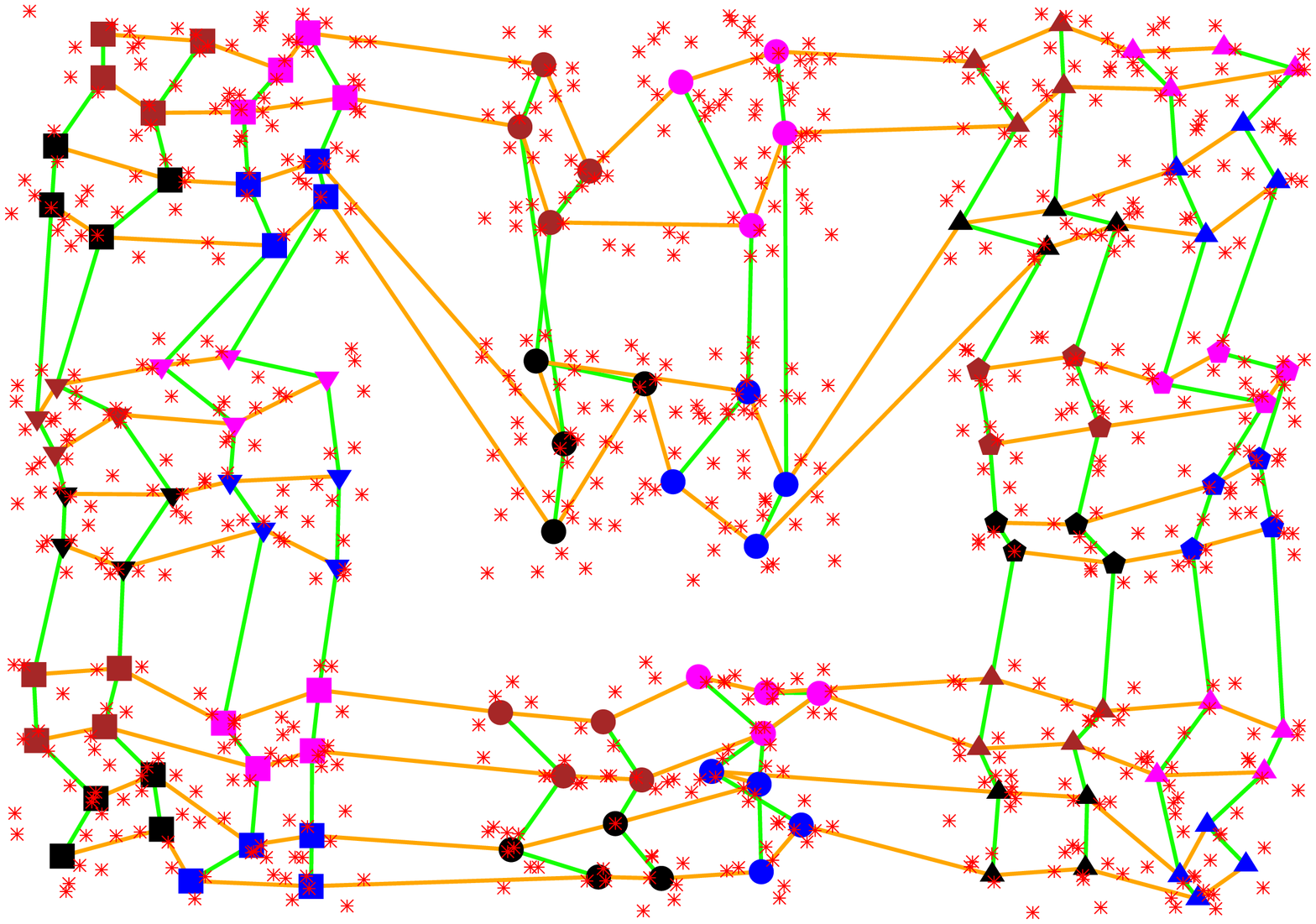}
\end{center}
\caption{\label{triscale-with-gap} Result of simulating a tri-scale metric SOM (metric $s_2$) with a ``stroke''-type gap in the middle of the map (that is, a group of map neurons is missing). The map is configured as a $3 \times 3$ array of groups, each of which is divided into a $2 \times 2$ array of subgroups, each of which is in turn $2 \times 2$ (so $12 \times 12$ overall). All map neurons lie within data clusters, and all subgroups lie within a single data cluster. This diagram indicates adjacency relationships between map nodes.}
\end{figure}

At this point, it becomes tempting to try a tri-scale metric version of the SOM algorithm. In the tri-scale simulation whose outcome is illustrated in Figure \ref{triscale-with-gap}, there is a $3 \times 3$ array of groups, each of which is divided into a $2 \times 2$ array of subgroups, each of which is in turn $2 \times 2$. All clusters contain map neurons, all map neurons lie within data clusters, and indeed all subgroups lie within a single data cluster. The top center group has split, with two subgroups in each of the top center data cluster and the middle center data cluster. In more detail, this approach required some tuning of the parameters $\mu$ and  $\lambda$ of the tri-scale metric, unlike the bi-scale metric, where the first value tried (namely $\lambda = 1$) worked. Simulation experiments showed that with $\mu = 6$, and $\lambda = 2$ or $3$ in second-stage training, 9 runs out of 10 achieved results where all map nodes were within clusters, and no subgroups were split between clusters. $\sigma(0)$ was set to the map radius in both stages in these simulations. While 10 out of 10 might seem better, it could also be argued that occasional failure qualitatively matches biological stroke recovery outcomes, where recovery may not occur or may not be full.


So the tri-scale approach works, but perhaps a simpler bi-scale approach would work too. So we conducted a simulation where the $12 \times 12$ map was divided into a bi-scale $6 \times 6$ array of $2 \times 2$ groups, the 4 central $2 \times 2$ groups were excised, corresponding to the single central $4 \times 4$ group in a standard bi-scale simulation, and all other simulation parameters kept the same - epochs, initial learning rate and neighborhood width, etc. This approach did not yield satisfactory results. Map nodes from 5 of the $2 \times 2$ groups migrated into the central cluster whose map units had been excised, with two of these groups being split between their original cluster and the new one, and one group ending up split between three clusters, with its 4th map node lying in the space between the four top-left clusters. This simulation is not illustrated here. In summary, the tri-scale approach worked, but the modified bi-scale approach did not.

\section{Simulations that vary learning rate}

\begin{figure}
\begin{center}
\includegraphics[width=12cm]{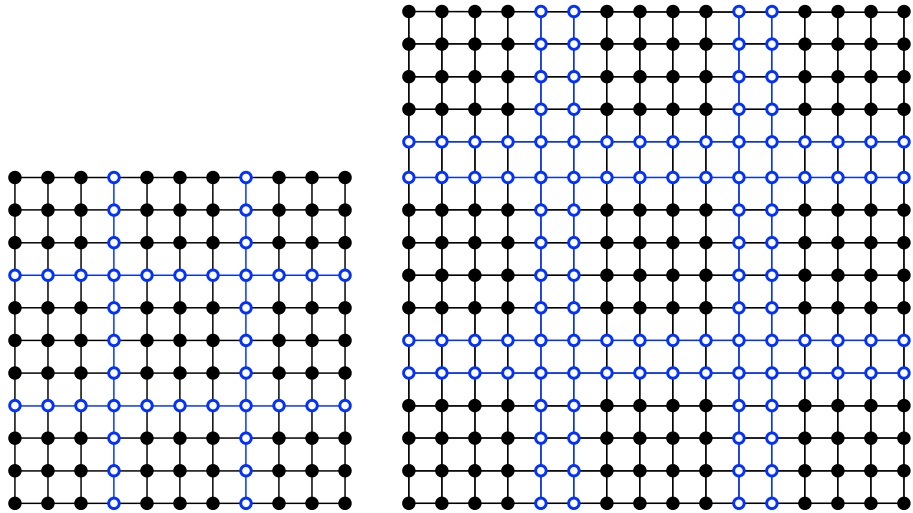}
\end{center}
\caption{\label{SARAbands} Topologies for SOMs with bands of neurons with a reduced learning rate. Such neurons are represented by open circles, and normal neurons are represented by closed circles. In the example on the left, every fourth row and column consists of map neurons with a lower learning rate. In the example on the right, there are four rows/columns of normal map neurons, then two rows/columns of lower-learning-rate neurons, and this pattern then repeats.}
\end{figure}

\begin{figure}
\begin{center}
\includegraphics[width=12cm]{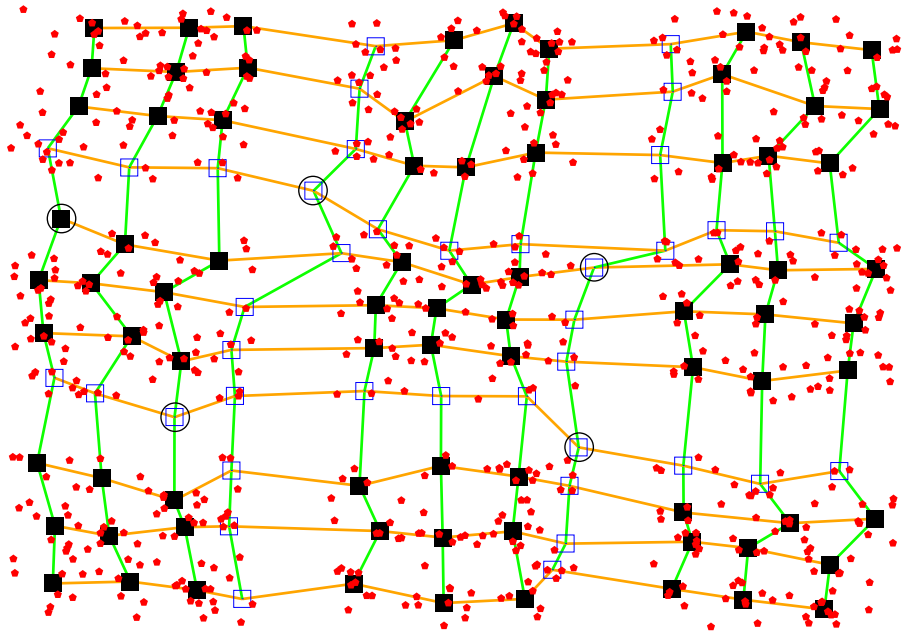}
\end{center}
\caption{\label{SARAbandsSim} Sample standard SOM with bands of neurons with a reduced learning rate, in which, as in Fig. \ref{SARAbands}, every fourth row and column (plotted as hollow squares) consists of map neurons with a lower learning rate. Circled neurons lie outside of data point clusters. The data consists of $3 \times 3$ clusters each of 80 pseudo-random points. In this simulation, the learning rate for SA map neurons was one fifth of the learning rate for RA neurons.}
\end{figure}

In \cite{KaasEtAl1981}, the term \emph{slowly adapting / SA} was used for neurons between areas of \emph{rapidly adapting / RA} neurons, see Fig. \ref{KaasFig8}. This terminology suggests an alternative simulation strategy in which the SOM map includes bands of neurons with a lower learning rate than standard neurons, as illustrated in Fig. \ref{SARAbands}. Also, \cite{KaasEtAl1981} do not spell out what particular role the SA neurons play in responding to the incoming sensory data in the case of a somatotopic map, and perhaps a simulation that incorporates low-learning-rate SA neurons might shed some light on that. However, in fact, maps produced by this variant of the SOM algorithm show isolated neurons between clusters of data points. As can be seen in Fig. \ref{SARAbandsSim}, which depicts the result of a simulation that uses the topology on the left side of Fig. \ref{SARAbands}, some of the isolated map neurons are the low learning rate (SA) neurons, and some are regular (RA) ones. No obvious pattern emerges from the maps that are generated, to allow an interpretation of the role of the SA neurons in responding to incoming data, and note that \cite{KaasEtAl1981} indicated that neurons activated by SA inputs were found at the margins between the neurons representing individual digits, whereas some of the simulated SA-related map neurons would in fact never `win' in the competitive phase for any stimulus in the training set, as other map neurons are closer to every training set member.

It might be argued that some refinement of this model - such as tuning the SA learning rate and/or using more SA rows and columns - could yield the desired network performance. The possible combinations of learning rate and numbers of SA rows/columns are next to endless, so such an argument is difficult to refute conclusively. However, simulations not diagrammed here showed that adding a second row and column of SA units (that is, the topology on the right side of Fig. \ref{SARAbands}) results in an even higher proportion of SA neurons lying outside the convex hulls of the clusters, and varying the learning rate to a tenth or two-fifths of the RA unit learning rate didn't improve the performance of the multiple learning rate model: SA and in some cases RA nodes lie outside clusters in these cases too.

So the bi-scale and tri-scale metric SOM variants, which do not specifically attempt to simulate SA neurons, appear to match the behavior of somatotopic maps better than SOMs with low-learning-rate bands. The question of the precise role of SA neurons in biological nets remains, in the sense that a resolution is not suggested by either model. The multiscale metric model doesn't have explicit neurons that correspond to the SA neurons, while the variable learning rate model does have ``SA'' map neurons but is problematical in having SA and RA map neurons between data clusters, and in having no clear ``behavioral'' role for the SA neurons in the map that emerges.


\section{Discussion and Conclusion}

\subsection{Summary of results}

The first modification of the SOM algorithm described in this paper, using the bi-scale or $s$-metric, is successful in preventing map neurons from falling between input clusters provided that the number and configuration of groups in the map is suitable for the clusters in the training data. This seems interesting in itself, and as already noted, it seems reasonable for a learning system whose gross input structure is predetermined, as is the case with the sensory inputs of the human body. From the point of view of modeling cortical maps, this means that map neurons will not correspond to places where there are no sensory neurons.

The second modification of the SOM algorithm, using the tri-scale or $s_2$-metric, is helpful in re-organizing simulated somatotopic and similar maps after simulated stroke damage.

The SA/RA phenomenon described earlier in relation to \cite{Sur1979, KaasEtAl1981} is modeled by the groups of neurons determined by the bi-scale metric (and by the tri-scale metric), in the sense that groups of map units correspond to regions of RA neurons in Sur, Kaas et al.'s  description. Although model neurons that might be designated as representing SA neurons do not explicitly appear in these models, we feel that this work provides a simple computational rationale for Sur, Kaas et al.'s observations of patterns of SA/RA neurons, and of the ability of the sensory cortex to reorganize when parts are no longer stimulated. The attempt at direct simulation of SA/RA neurons using a SOM variant that has neurons with different learning rates did not give a good match with cortical map reorganization performance.

The bi-scale and tri-scale metric computational models do not provide direct insight into cortical wiring, though they suggest that a wiring model consistent with these metrics may have something going for it.

The bi-scale metric method appears to work by changing the ``pull'' of neighboring neurons during the SOM adaptive process to favor map neurons in the same group as a particular map neuron. Considering a map neuron $A$ whose weight vector currently lies between clusters, when an adjacent map neuron $B$ in the same group is the winner in the competitive process, the cooperative and adaptive processes result in a greater change to the weight vector for $A$ than is the case if the winning neuron is another adjacent map neuron $C$ that lies in a different group. Thus $A$'s weight vector is drawn towards that for $B$ over time. So if $B$'s weight vector lies within a cluster of data points, $A$'s weight vector will move into that cluster, too, and will then have the opportunity to be a winning node itself. In effect, nodes that are in the same group stick together, so that a `straggler' cluster-buddy neuron will be pulled into a cluster of data points by the algorithm.

When there is a natural correspondence between the number of groups of map neurons defined by the bi-scale metric and the number of input data clusters, this process seems to work perfectly in the cases examined. When there are more groups of map neurons than clusters, the spare groups cover more than one cluster, either by having, say, two groups in one cluster, or, as in the case of the three groups covering the 2 ``thumb'' clusters in Fig. \ref{Hand15x9standardSOMe=5K} where two clusters are close together, by splitting a group across two clusters.

In the case where there are fewer groups than clusters, experiments not diagrammed in this paper
show that groups may split across clusters and, for this reason, there may be map neurons between clusters. In other words, the bi-scale metric method relies on there being a sufficient number of clusters suitably arranged.

For biological somatotopic map organization, this means that the organism would have to know at the time of neural development how many body regions there are, and which ones are adjacent. This is a much simpler problem than that of individually mapping every sensory neuron to a suitable location in a somatotopic map region. Body structure is consistent within a species, barring developmental variations like supernumerary digits, and indeed is similar between related species, and this presumably makes it easier for evolutionary processes to incorporate knowledge of body structure into neural wiring.

\subsection{Further issues}

One outstanding issue is the problem of having a biologically plausible and effective version of the SOM algorithm that performs continuous learning (cortical map plasticity). The problem is that decreasing neighborhood width with time (or the parallel narrowing of neighborhood sets in the neighborhood set version of the algorithm) largely freezes the learning process. In terms of direct modifications of the SOM algorithm, there seem to be two possibilities: either there could be something analogous to resetting the neighborhood width to a larger value and rerunning the learning algorithm, or else there could be a substantially different algorithm that provides continuous updating of cortical maps.

Dynamic or evolving algorithms related to the SOM algorithm have been developed: they tend to involve dealing with new types of input by adding map neurons, as in ESOM \cite{DengKasabov2003}, an interesting algorithm that starts with an empty map and adds map neurons when the current input is not sufficiently close to any of the current map neurons. The notion of ``sufficiently close'' is determined by a distance threshold parameter $\varepsilon$, and they use a constant learning rate. They report that this algorithm works well for the range of applications and datasets that they were interested in. Biological realism was probably not among their goals. An issue for this and all learning algorithms, if applied to cortical map modeling, is differing sensory sensitivity, such as fingertips versus back, or fovea versus other retinal regions. An ESOM-based model might deal with this by using differing values of $\varepsilon$ for different parts of input spaces. In a bi-scale metric model this issue comes down to the pre-defined network topology.  Another issue for biologically plausible modeling using ESOM might be the reliance on the availability of unused neurons in the right place, or else neurogenesis \cite{Gould1999} in the right place at the right time. However, if the actual plasticity mechanism supports continuous, on-line, updating of cortical maps, then an algorithm similar to this one might be a promising way forward.

One essence of the SOM variants described in this paper is that they introduce a hierarchical , pre-specified structure: using the bi-scale metric produces a two-level hierarchy (the whole map and the subgroups), and using the tri-scale metric produces a three-level hierarchy. In this respect it resembles algorithms such as the Growing Hierachical Neural Gas (GHNG) algorithm of \cite{PalomoLopez2017}.  This performs well on their tasks and appears to have the potential to match any hierarchical structure. It differs from the algorithm reported in this paper in apparently not being biologically inspired nor biologically applicable.

If resetting and relearning with a SOM-like algorithm is the right choice, then it would be necessary to present training data to the learning algorithm in a way that includes (a subset or summary of) historical stimuli as well as recent, different stimuli, when those historical stimuli are not actually currently occurring. One might speculate that sleep, and/or dreaming (which occurs in a wide range of vertebrate brains),
is, among other things, an opportunity for this sort of off-line relearning. At such times, the brain's motor system is mostly inactive/offline (or does not effect actual muscular action), and adjustment of cortical connections would thus be more feasible. Some of the sensory hallucinations that occur in dreams might serve as a summary of historical and recent stimuli for training. Illustrations in recent work of \cite{SiclariEtAl2017} seem to indicate that the brain areas where some cortical maps occur are among those that are active during dreaming. If, however, the actual plasticity mechanism supports continuous, on-line, updating of cortical maps, then it seems that a quite different algorithm might be required. 

The results reported in this paper, combined with results from the literature, allow us to raise some questions about the way cortical maps are created and updated. What stimuli are used for maintaining and updating cortical maps? The fact that map territory is re-allocated in the case of amputation suggests that continuous (re-)training may be necessary. This could be coming from environmental patterns, or from endogenously generated data during sleep, or a mixture of these. Then new input stimulus patterns could simply be integrated into the stream of material that is being used to maintain and update the cortical maps. In the case of visceral (e.g. \cite{JensenEtAl2013}) and proprioceptive sensation, sensory input is received regularly, so there is a clear mechanism for map maintenance.

\subsection{Further possibilities}

The effectiveness of the bi-scale and tri-scale metrics described in this paper suggests that it may be fruitful to experiment with other non-standard metrics. As mentioned above, experiments by the author reported elsewhere \cite{Wilson2017} showed that the $\ell^p$-norm-based metrics ($p \ge 1$) and the max-norm-based metric appear to work when used in the cooperative phase of the SOM algorithm, while others do not. It is easy to imagine bi-scale and tri-scale style metrics that group the map nodes in ways other than the square groupings used in this paper, and such metrics might be useful with particular data distributions. 

The metric perspective may also shed light on the significance of connectivity at the boundaries between regions of cortex responding to specific sensory receptor groups, for example in mouse barrel cortex, an area that has been extensively investigated both empirically and with computational models (e.g. \cite{SaridEtAl2015} and references therein). Higher-dimensional analogs of the metrics used in this paper might be fruitful (e.g. 3-D maps with modified metrics). It is natural to map skin sensation onto a 2-D (cortical) map, but 3-D visual sensation might benefit from a 3-D map, though this might be more helpful for robot vision than for biological vision, which seems to be committed to 2-D.

It might be worth trying a 3-dimensional analog of the bi-scale and/or tri-scale metrics for modeling neural systems with no natural two dimensional expression, such as sensory and effector neural systems in organs like the liver \cite{JensenEtAl2013}, and for that matter a tube topology analog of the 2-dimensional version for intestinal sensory and effector systems.

The problems motivating the work reported in the current paper come from modeling biological neural systems (at a fairly coarse level). There may also be applications of bi-scale/tri-scale metrics or bespoke metrics in other application areas for SOMs. Such bespoke metrics may already exist in corners of mathematics, just waiting for an application to be matched to them.

While the bi-scale-metric method does not directly translate to a biological neural mechanism, perhaps it may suggest something to neuroscientists interested in the neuroanatomy and mechanisms hinted at by the research summarised in Fig. \ref{KaasFig8}.


%


\bibliography{mybibfile}

\end{document}